\documentclass[letterpaper, 10 pt, journal, twoside]{IEEEtran}

\IEEEoverridecommandlockouts                              

\usepackage[T1]{fontenc}

\usepackage{tikz} 

\usepackage{amsfonts}   
\usepackage{amsmath}    
\usepackage{amssymb}    
\usepackage{siunitx}
\usepackage{pifont}   
\usepackage{blkarray} 
\usepackage{bigstrut} 

\usepackage{booktabs}
\usepackage{makecell}  
\usepackage[flushleft]{threeparttable}  
\usepackage{multirow}
\usepackage{rotating}

\usepackage[compact]{titlesec}  
\usepackage[inline]{enumitem} 
\usepackage{xspace}    
\usepackage{xcolor}    
\usepackage{textcomp}  
\usepackage{gensymb}   
\usepackage{graphicx} 
\usepackage[caption=false,font=footnotesize]{subfig}
\usepackage[export]{adjustbox}  

\usepackage[pdfencoding=auto, hidelinks]{hyperref} 
\usepackage[hang,flushmargin]{footmisc}

\titlespacing{\section}{0pt}{*1}{*.5}
\titlespacing{\subsection}{0pt}{*.5}{*0}

\newcommand{\cmark}{\ding{51}}  

\newcommand{\asterisk}{\ding{83}}

\newcommand{\refeqn}[1]{Eq.~\ref{#1}}
\newcommand{\reffig}[1]{Fig.~\ref{#1}}
\newcommand{\refsec}[1]{Sec.~\ref{#1}}

\newcommand{\reftab}[1]{Table~\ref{#1}}

\begin{document}

\title{PADLoC: LiDAR-Based Deep Loop Closure Detection and Registration Using Panoptic Attention}

\author{
José Arce$^{1}$, 
Niclas Vödisch$^{1}$, 
Daniele Cattaneo$^{1}$, 
Wolfram Burgard$^{2}$, 
and Abhinav Valada$^{1}$
\thanks{© 2023 IEEE. Personal use of this material is permitted. Permission from IEEE must be obtained for all other uses, in any current or future media, including reprinting/republishing this material for advertising or promotional purposes, creating new collective works, for resale or redistribution to servers or lists, or reuse of any copyrighted component of this work in other works.}
\thanks{This work was funded by the European Union’s Horizon 2020 research and innovation program under grant agreement No 871449-OpenDR and the DFG Emmy Noether Program.}%
\thanks{$^{1}$ José Arce, Niclas Vödisch, Daniele Cattaneo, and Abhinav Valada are with the Department of Computer Science, University of Freiburg, Germany.}%
\thanks{$^{2}$ Wolfram Burgard is with the Department of Engineering, University of Technology Nuremberg, Germany.}%
\thanks{Digital Object Identifier (DOI): 10.1109/LRA.2023.3239312}
}

\markboth{T\MakeLowercase{his paper appeared in:} IEEE ROBOTICS AND AUTOMATION LETTERS, VOL. 8, ISSUE 3, MARCH 2023}%
{Arce \MakeLowercase{\textit{et al.}}: PADLoC: LiDAR-Based Deep Loop Closure Detection and Registration Using Panoptic Attention}

\maketitle


\begin{abstract}
    A key component of graph-based SLAM systems is the ability to detect loop closures in a trajectory to reduce the drift accumulated over time from the odometry. Most LiDAR-based methods achieve this goal by using only the geometric information, disregarding the semantics of the scene. In this work, we introduce PADLoC for joint loop closure detection and registration in LiDAR-based SLAM frameworks. We propose a novel transformer-based head for point cloud matching and registration, and to leverage panoptic information during training time. In particular, we propose a novel loss function that reframes the matching problem as a classification task for the semantic labels and as a graph connectivity assignment for the instance labels. During inference, PADLoC does not require panoptic annotations, making it more versatile than other methods. Additionally, we show that using two shared matching and registration heads with their source and target inputs swapped increases the overall performance by enforcing forward-backward consistency. We perform extensive evaluations of PADLoC on multiple real-world datasets demonstrating that it achieves state-of-the-art results. The code of our work is publicly available at \url{http://padloc.cs.uni-freiburg.de}.

\end{abstract}




\section{Introduction}

\begin{figure}[t]
    \centering
    \includegraphics[width=1\linewidth]{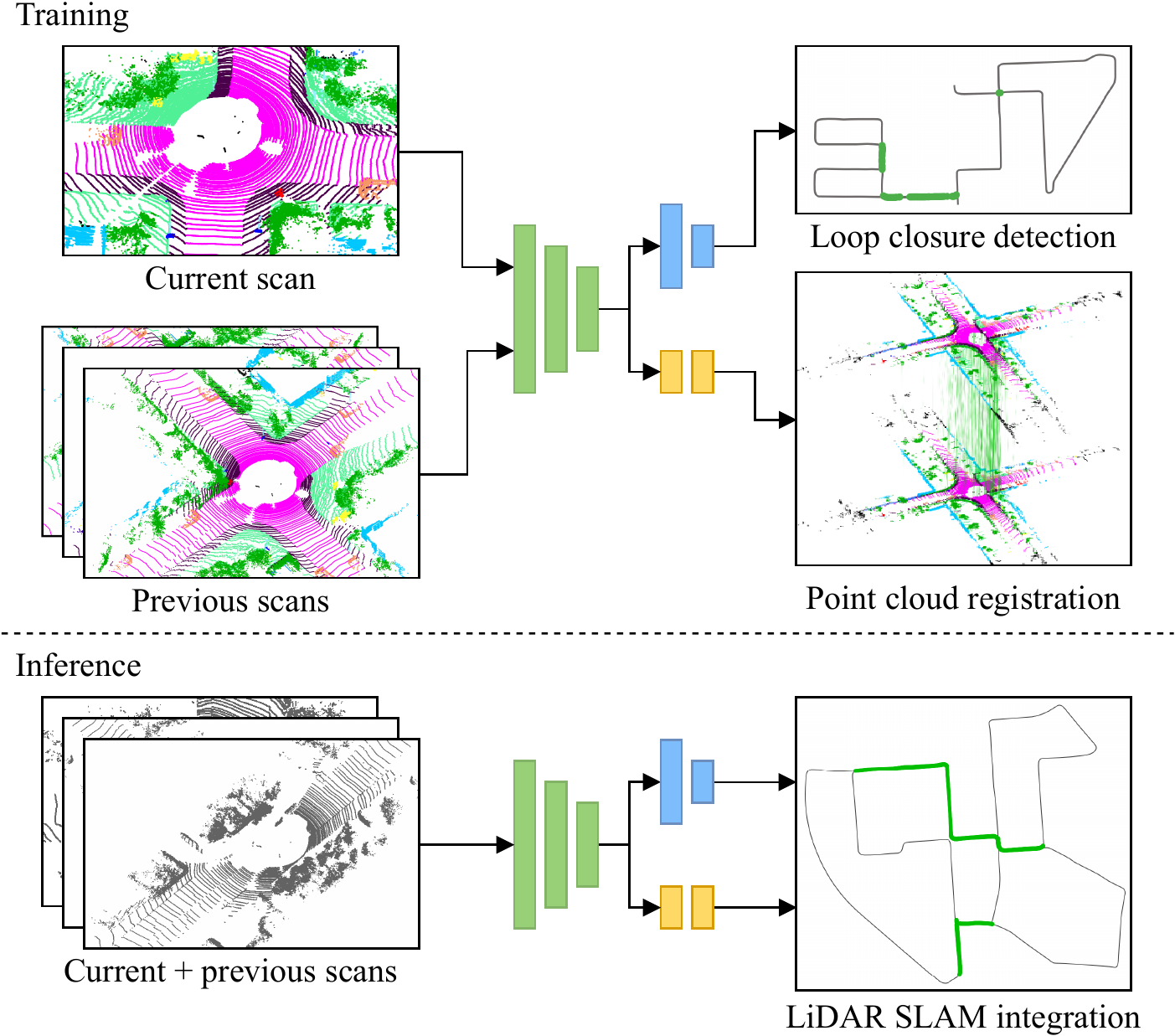}
    \vspace{-.7cm}
    \caption{We propose PADLoC for joint loop closure detection (green areas on the map) and point cloud registration in LiDAR-based SLAM. In addition to geometric information, we leverage panoptic segmentation annotations during training to facilitate more robust point matching. During inference, PADLoC does not require any panoptic information.}
    \label{fig:teaser}
    \vspace{-.6cm}
\end{figure}

\IEEEPARstart{S}{imultaneous} Localization and Mapping (SLAM) is a core task of autonomous mobile robots. Typically, SLAM approaches consist of two steps: alignment of consecutive measurements, e.g., from wheel odometry, followed by loop closure detection and registration. Reliable loop closure detection enables a robot to recognize places it has seen before to optimize its world representation and belief of its current position, reducing the drift over time. Thus, it is considered a fundamental component of SLAM systems. Many SLAM systems have been proposed for different sensor modalities including cameras~\cite{voedisch2022continual} and LiDARs~\cite{li2021saloam}. While vision-based methods fail in challenging lighting conditions such as illumination changes, LiDAR-based approaches are more robust to such alterations and provide a more accurate representation of the environment. In this work, we address the joint problem of loop closure detection and map registration for LiDAR-based SLAM.  A high-level overview of our approach is depicted in~\reffig{fig:teaser}.

Similar to other fields, learning-based approaches have started to replace handcrafted methods~\cite{bevsic2022dynamic, gosala2022bird}. Typically, deep neural networks predict point correspondences which are then used in differential singular value decomposition (SVD) to compute the transformation between two point clouds~\cite{cattaneo2022lcdnet, wang2019deep}. Motivated by the success of transformers in natural language processing and computer vision tasks, attention-based architectures were recently introduced for point cloud registration~\cite{wang2019deep, qin2022geometric, yew2022regtr} to encode context across points. While existing works do not consider the semantic meaning of the different inputs to a transformer cell, i.e., queries, keys, and values, we explicitly take advantage of the internal structure by feeding in abstract features and raw points separately.

Although geometric information suffices for classical point cloud registration such as Iterative Closest Point (ICP)~\cite{zhang1994iterative}, they can be further stabilized by integrating semantic information~\cite{li2021saloam, chen2019sumapp, kong2020semantic}. Inspired by recent semantic mapping approaches~\cite{chen2019sumapp, radwan2018vlocnet++} and  methods that exploit panoptic information for vision-based loop closure detection~\cite{yuan2021svloop}, we leverage panoptic segmentation of point clouds in this work. Unlike related methods, our approach requires panoptic labels only while training but not during deployment, making it more versatile. We evaluate the loop closure detection and point cloud registration performance on three real-world autonomous driving datasets, namely, KITTI~\cite{geiger2012are}, Ford campus~\cite{pandey2011ford}, and an in-house dataset recorded in Freiburg, Germany. We compare against both state-of-the-art handcrafted and deep learning-based methods and demonstrate that PADLoC achieves state-of-the-art performance. We also present several ablation studies on the different components of our approach validating our architectural design choices.

The main contributions of this work are as follows:
\begin{enumerate}[label={\arabic*)},topsep=0pt]
    \item We propose PADLoC, a transformer encoder architecture for point cloud matching and registration. Unlike existing methods, we use separate inputs as keys, values, and queries effectively, exploiting the transformer structure.
    \item We define a novel loss function that leverages panoptic information for registration. We further propose formulating both geometric and panoptic registration losses as bidirectional functions that greatly improve performance.
    \item We study the effect of multiple weighting methods in SVD to enhance point matching.
    \item We extensively evaluate our proposed approach and compare it to other point cloud matching and registration methods, using two openly available datasets and \mbox{in-house} data recorded in Freiburg, Germany.
    \item We release our code and the trained models at \mbox{\url{http://padloc.cs.uni-freiburg.de}}.
\end{enumerate}

\section{Related Work}

In this section, we first provide an overview of LiDAR-based loop closure detection techniques for SLAM, followed by various methods for point cloud registration, and finally describe approaches that leverage semantic segmentation for either task.


{\parskip=3pt
\noindent\textit{Loop Closure Detection:} 
Traditionally, handcrafted methods for LiDAR loop closure detection can be categorized into local feature-based and global feature-based methods. Inspired by the success of local feature-based methods in images, approaches from the first category design similar descriptors and adapt them to 3D point cloud data. 3D keypoint descriptors such as Fast Point Feature Histograms (FPFH)~\cite{rusu2009fast} and Normal-Aligned Radial Features (NARF)~\cite{steder2011narf} are used to extract local features, which are then aggregated in a bag-of-word model to detect loop closures. More recently, HOPN~\cite{Lun2022} exploits a bird's-eye view (BEV) representation and normal information to increase robustness to noise and viewpoint changes. Global feature-based approaches, on the other hand, summarize the whole point cloud into a single fingerprint, which is then compared against the fingerprints from past frames to detect loops. The M2DP~\cite{He2016} descriptor projects the point cloud into multiple 2D planes and combines density information computed on each plane into a global descriptor. Scan Context~\cite{giseop2018scan} combines a polar coordinate representation with partitioning to generate an image as a global descriptor. Subsequent works extended this method by adding additional information such as intensity~\cite{wang2020intensity} and semantic data~\cite{li2021ssc}. Recently, many deep learning-based approaches have been proposed to overcome some of the limitations of handcrafted methods. PointNetVLAD~\cite{Uy_2018_CVPR} is built on top of the PointNet~\cite{Qi_2017_CVPR} architecture and generates a compact descriptor. OverlapNet~\cite{chen2020overlapnet} projects the point cloud into a range image and predicts the overlap and the yaw misalignment between a pair of frames. To increase viewpoint robustness and to reduce inference time, OverlapTransformer~\cite{Junyi2022} adapts OverlapNet by including a transformer module. In this work, we build upon LCDNet~\cite{cattaneo2022lcdnet} that uses learning-based feature extraction to generate global descriptors. LCDNet significantly improves loop closure in challenging conditions, such as reverse loops and, unlike other methods, does not require an ad-hoc function to compare two global descriptors.
}


{\parskip=3pt
\noindent\textit{Point Cloud Registration:}
Standard techniques for point cloud registration can be broadly classified into two main categories. The first category comprises the Iterative Closest Point (ICP) algorithm~\cite{zhang1994iterative} and its variants~\cite{chen2019sumapp,bouaziz2013sparse}. These methods require an initial guess on the transformation and then iteratively alternate between finding matches between points by exploiting some heuristics and estimating the transformation based on these matches. Methods of the second category use a two-stage approach. They first extract local point features, e.g., FPFH~\cite{rusu2009fast}, and then regress the transformation using robust estimators such as RANSAC~\cite{fischler1981random}. While methods of the first category are prone to get stuck in local minima if the provided initial guess is not accurate enough, approaches of the second category are sensitive to noise and incorrect matches. Many deep learning-based approaches have also been proposed to solve the point cloud registration task. PointNetLK~\cite{Aoki_2019_CVPR} is a pioneering work that combines an architecture inspired by PointNet~\cite{Qi_2017_CVPR} and a modified Lucas-Kanade algorithm to iteratively improve the registration. Inspired by the success of transformers in other fields, Deep Closest Point~\cite{wang2019deep} uses an attention-based module to predict soft matches between two point clouds, which are fed to a differentiable SVD layer to infer a rigid transformation. Following the same idea, both GeoTransformer~\cite{qin2022geometric} and REGTR~\cite{yew2022regtr} directly learn to predict point correspondences using both self and cross-attention. Our previous work LCDNet~\cite{cattaneo2022lcdnet} combines a state-of-the-art feature extraction architecture with a place recognition head and a relative pose head for simultaneous loop closure detection and point cloud registration. In this work, we adapt LCDNet~\cite{cattaneo2022lcdnet} by integrating a transformer-based registration and matching module.
}


\begin{figure*}[t]
    \centering
    \includegraphics[width=.95\linewidth]{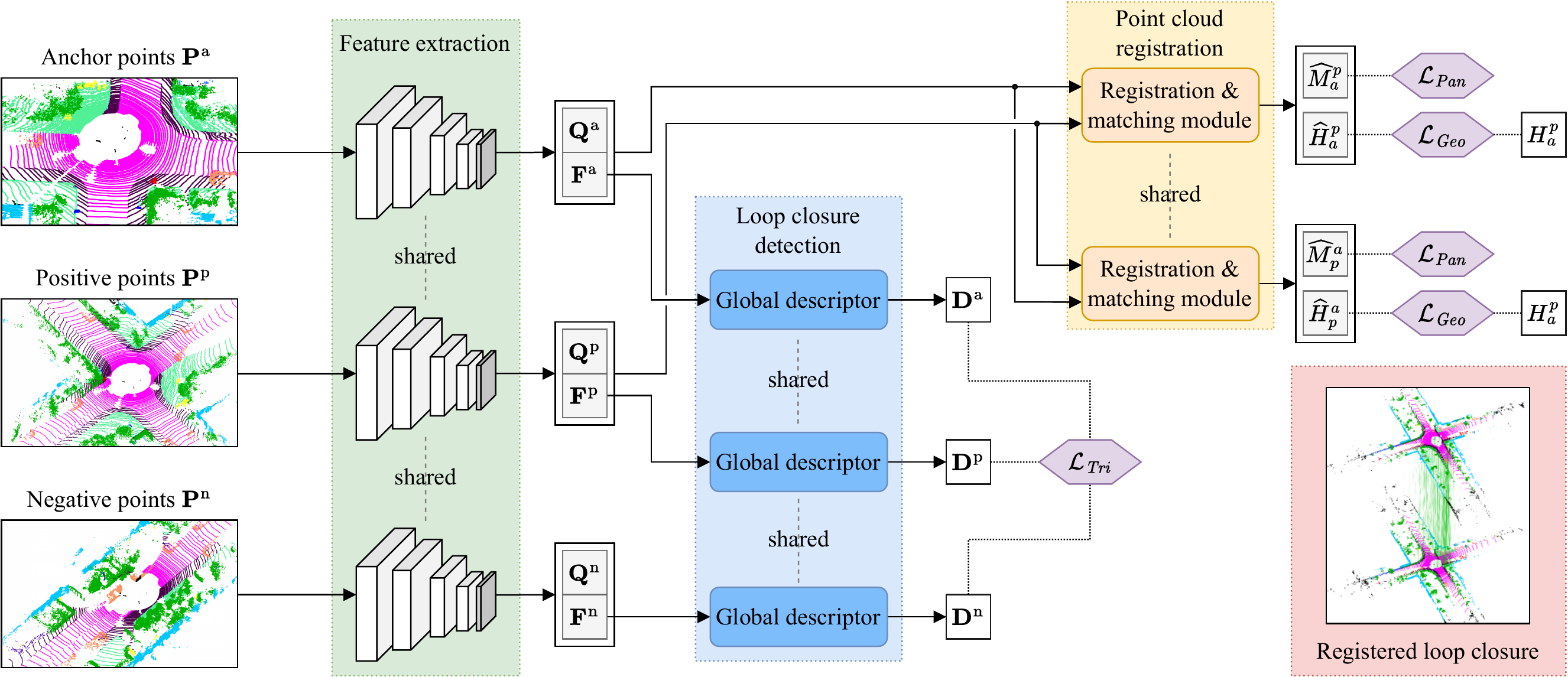}
    \vspace*{-.2cm}
    \caption{Overview of our proposed PADLoC architecture for joint loop closure detection and point cloud registration. It consists of a shared feature extractor (\textcolor[HTML]{82B366}{green}) followed by a global descriptor head (\textcolor[HTML]{6C8EBF}{blue}) for loop closure detection and a registration and matching module (\textcolor[HTML]{D79B00}{orange}) to estimate the 6-DoF transform between two point clouds (\textcolor[HTML]{B85450}{red}). To train the global descriptor, we use a triplet loss (\textcolor[HTML]{9673A6}{purple}) that compares the anchor point cloud with a positive and negative sample. For training the registration module, we leverage losses (\textcolor[HTML]{9673A6}{purple}) based on both geometric and panoptic information. Note that during inference, no panoptic annotations are required, making PADLoC more versatile than other methods.}
    \label{fig:overview}
    \vspace*{-.5cm}
\end{figure*}

{\parskip=3pt
\noindent\textit{Semantic-Aided Mapping and Localization:}
Only a handful of works have proposed to leverage semantic information for large-scale mapping and localization~\cite{chen2019sumapp,ballardini2019}, and particularly for loop closure detection. Based on semantic segmentation, SuMa++~\cite{chen2019sumapp} filters dynamic objects from a LiDAR-based map and extends the ICP algorithm with additional semantic constraints. While SuMa++ does not utilize semantic information for loop closure detection, RINet~\cite{li2022rinet} explicitly addresses LiDAR-based place recognition via a rotation invariant global descriptor combining semantic and geometric information. For the same task, SGPR~\cite{kong2020semantic} builds a graph representation of point clouds, which are enriched by both semantic and instance segmentation and perform graph similarity matching. \mbox{SA-LOAM}~\cite{li2021saloam} integrates a semantic-aided variant of ICP into the popular LOAM pipeline for point cloud registration. To address loop closure, it uses a similar graph representation as Kong~\textit{et~al.}~\cite{kong2020semantic}. SV-Loop~\cite{yuan2021svloop} is a loop closure detection method for vision-based SLAM. It separately proposes loop closure candidates based on raw images and panoptic segmentation maps, which are then fused to extract the most feasible candidates. In our approach, we exploit panoptic annotations of point clouds while predicting both loop closure detection and point cloud registration. Additionally, we only utilize them during the training process but not for deployment, making the method more versatile.
}

\section{Technical Approach}

In this section, we introduce our novel PADLoC architecture for joint loop closure detection and point cloud registration. First, we detail the overall approach comprising the modules shown in \reffig{fig:overview}. We then describe the loss functions that we employ, including our proposed loss that leverages panoptic annotations of point clouds.


\subsection{Model Architecture}

\begin{figure}[t]
    \centering
    \includegraphics[width=.95\linewidth]{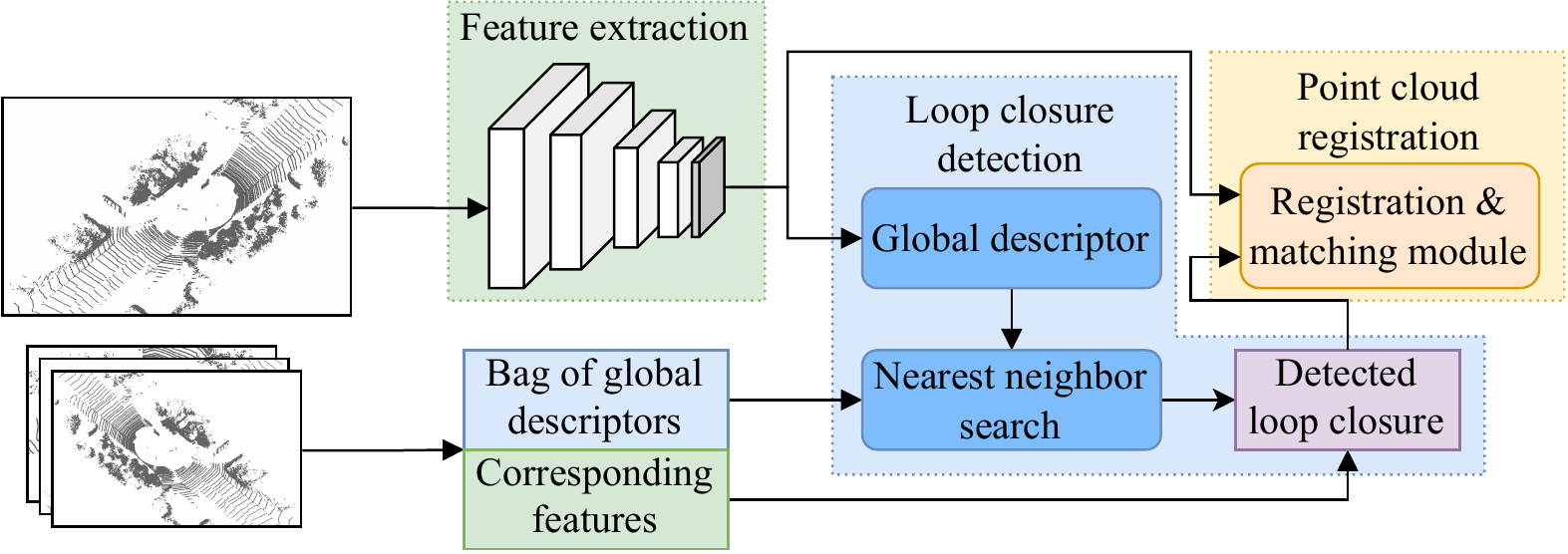}
    \vspace*{-.2cm}
    \caption{During inference, PADLoC does not require panoptic annotations to extract features. To detect a loop closure, we perform a nearest neighbor search in the global descriptor space. If a loop is found, the 3D transformation is computed using the registration and matching module.}
    \label{fig:inference}
    \vspace*{-.5cm}
\end{figure}

In this section, we describe the individual components of the PADLoC architecture. We build upon our previously proposed LCDNet~\cite{cattaneo2022lcdnet}, where instead of using a differentiable approximation of the optimal transport to obtain point matches, we propose to leverage the cross-attention matrices of transformers. The learnable keys, queries, and values weights yield a better latent representation of the features, and thus more reliable matches. As depicted in \reffig{fig:overview}, the overall PADLoC architecture consists of three modules: feature extraction, loop closure detection, and point cloud registration. During training, we employ a triplet-based training scheme by feeding in an anchor point cloud along with a positive sample of a loop closure and a negative sample.
Unlike other methods and as shown in \reffig{fig:inference}, PADLoC does not require panoptic annotations during inference.


{\parskip=3pt
\noindent\textit{Feature Extraction:}
The feature extraction backbone converts raw input scans into a high-dimensional embedding that is used as a common input for both loop closure detection and point cloud registration. It effectively exploits global and local contexts and is built upon the PV-RCNN architecture~\cite{shi2020pvrcnn}. In detail, a point cloud $\mathbf{P}$, comprising 3D coordinates and reflectance values, is discretized into a voxel grid which is then passed through four sparse 3D convolutional layers to generate the feature maps at different resolutions. The final feature map is then stacked to form a BEV feature map. Additionally, the original point cloud is downsampled using the Farthest Point Sampling (FPS) algorithm to uniformly select $n$ keypoints. The feature vector of each sampled keypoint is assembled by combining the feature maps from each convolutional layer in a neighborhood of the sampled keypoint using the Voxel Set Abstraction module~\cite{shi2020pvrcnn}. The raw input of each sampled keypoint is also appended to each feature vector, along with the corresponding entry in the BEV feature map. Finally, these intermediate features are fed through a multilayer perceptron to obtain the final feature vector for each sampled point. This module thus outputs the sampled keypoints $\mathbf{Q}$ and the corresponding features $\mathbf{F}$.
}


{\parskip=3pt
\noindent\textit{Loop Closure Detection:}
The global descriptor module of PADLoC further encodes the previously extracted features to perform loop closure detection. For this task, we employ the NetVLAD layer~\cite{arandjelovic2018netvlad} to convert the feature vectors $\mathbf{F}$ of the anchor, the positive, and the negative points to their respective final descriptor $\mathbf{D}$. In detail, NetVLAD learns $k$ clusters along with corresponding descriptors, which are aggregated in a single descriptor $v$ for the entire point cloud. The final descriptors~$\mathbf{D}$ of length $g$ are then obtained via a context gating layer. This learnable pooling operation with weights $\mathbf{W}_G$ and bias $\mathbf{b}_G$ is defined as
\begin{equation}
    \mathbf{D} = \sigma \big( \mathbf{W}_G \cdot v + \mathbf{b}_G \big) \odot v,
\end{equation}
where $\sigma(\cdot)$ refers to the logistic sigmoid function and $\odot$ denotes the element-wise multiplication.

During inference, the descriptors are stored in such a manner that allows for efficient querying of the nearest neighbor in descriptor space. If the distance between the descriptor of the current scan and its nearest neighbor is below a predefined threshold, they are considered to form a loop closure. To avoid matching consecutive scans, we introduce a small temporal distance between the current scan and potential neighbors. 
}


\begin{figure}[t]
    \centering
    \includegraphics[width=.95\linewidth]{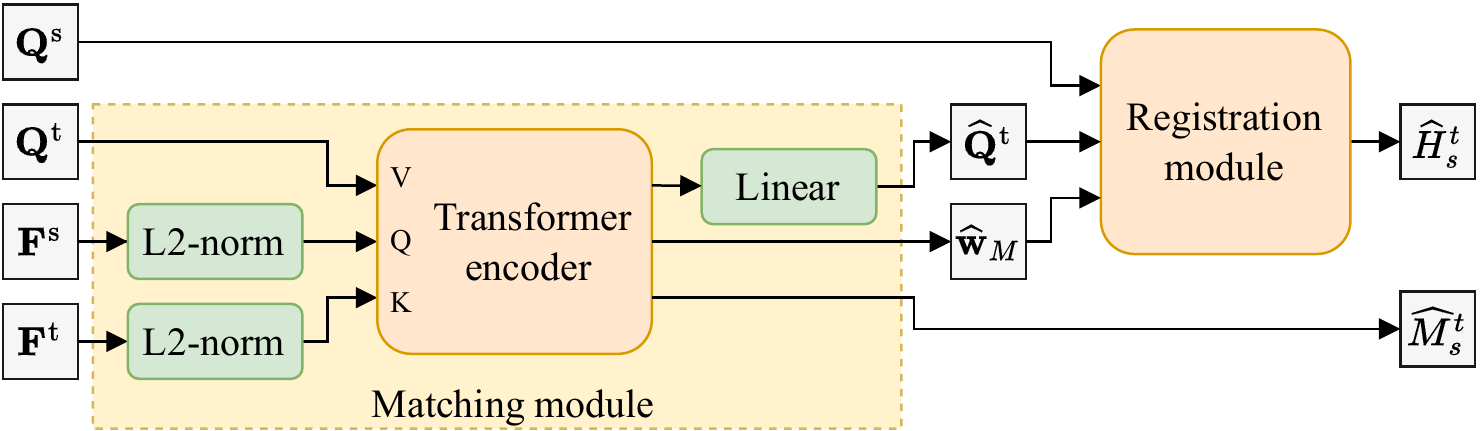}
    \vspace{-.2cm}
    \caption{The matching module consists of a transformer encoder that takes the extracted features of the source keypoints $\mathbf{F}^s$ as query, the features of the target keypoints $\mathbf{F}^t$ as key, and the corresponding target keypoints $\mathbf{Q}^t$ as value. It outputs both soft correspondences $\widehat{M}^t_s$ and projected target points $\widehat{\mathbf{Q}}^t$ along with confidence weights $\widehat{w}_M$. The latter is fed together with the source keypoints $\mathbf{Q}^s$ to a registration module that performs weighted SVD to estimate the final transform $\widehat{H}_s^t$.}
    \label{fig:registration_module}
    \vspace{-.5cm}
\end{figure}

{\parskip=3pt
\noindent\textit{Point Matching:}
\label{sec:matching}
The matching module shown in \reffig{fig:registration_module} predicts soft correspondences \(\widehat{M}_s^t\) between keypoints $\mathbf{Q}^s$ and $\mathbf{Q}^t$ of a source point cloud \(s\) and a target point cloud \(t\), respectively. Additionally, it outputs projected target coordinates \(\widehat{\mathbf{Q}}^t\) which are linear combinations of the original target coordinates with a one-to-one pairing with the points of the source set and a confidence weight \(\widehat{w}_M\) for each of these matches. Inspired by the success of transformers in related tasks, we propose a novel architecture that performs cross-attention directly on the encoder part, obviating the need for a decoder by feeding independent inputs for the queries, keys, and values.
\begin{equation}
    \widehat{\mathbf{Q}}^t = \mathbf{W}_Q \cdot \operatorname{TEL} \big( \mathbf{F}^s, \mathbf{F}^t, \mathbf{Q}^t \big) + \mathbf{b}_Q,
\end{equation}
where \(\operatorname{TEL} (q, k, v)\) is a transformer encoder layer, as defined in \cite{vaswani2017attention}, but applied to independent query \(q\), key \(k\), and value \(v\) inputs. \(\mathbf{W}_Q \in \mathbb{R}^{3\times f}\) and \(\mathbf{b}_Q \in \mathbb{R}^3\) are learnable weights and biases used to reduce the dimensionality of the output from the size \(f\) of the features $\mathbf{F}$ to 3D space. We directly use the encoder's attention matrix as our matching \(\widehat{M}_s^t\), since it already encodes the similarity between the features of the two sets of points. The output of the transformer encoder is given by the matrix product of the attention matrix and the weighted values input. Since in our case, we supply the target coordinates as the value input, it follows that the output of the transformer encoder corresponds to linear combinations of the input target points, weighted by the attention matrix and denoted by \(\widehat{\mathbf{Q}}^t\). These projected points have a one-to-one correspondence with those of the anchor point cloud. Moreover, each row in the attention matrix represents the probability distribution of matching the corresponding point from the source set to all of the points from the target set, given that it is non-negative and adds up to one due to the use of the softmax function.

From the matching matrix \(\widehat{M}_s^t\), we compute a confidence weight for every pair of point correspondences by penalizing the dispersion of the distributions represented by each row. We propose using a diversity metric for that purpose, such as the Shannon Entropy ($\operatorname{E}$), the order-\(r\) Hill number ($\operatorname{D}^r$), or the Berger-Parker index ($\operatorname{BP}$), defined as
\begin{align}
    \operatorname{E} (\mathbf{p}) &= - \sum_i p_i \cdot \log (p_i) \label{eq:shannon},\\
    \operatorname{D}^r (\mathbf{p}) &= \Big( \sum_i p_i^r \Big) ^ {\frac{1}{1 - r}}, \label{eq:hill}\\
    \operatorname{BP} (\mathbf{p}) &= \max (\mathbf{p}) \label{eq:berger},
\end{align}
where \(\mathbf{p}\) is a vector of probabilities.

The weights \(\widehat{w}_M\) are obtained using either of the aforementioned metrics by normalizing their output to a \([0, 1]\) range, where the two extreme weights of 0 and 1 respectively correspond to a uniform and an infinitely sharp distribution.
}


{\parskip=3pt
\noindent\textit{Point Cloud Registration:}
To obtain the final relative transformation $\widehat{H}_s^t$ from a source point cloud to a target point cloud, we perform a weighted version of the Kabsch-Umeyama algorithm that finds the optimal translation and rotation between two sets of points by minimizing the root mean square error of the point pairs. First, the correspondences between the sampled source keypoints $Q^s$ and the projected target keypoints $\widehat{Q}^t$ are weighted by the matching confidences $\widehat{w}_M$. Subsequently, the optimal translation is computed as the difference between the weighted centroids of the two point clouds. Finally, the optimal rotation is obtained via SVD of the weighted covariance matrix of the two sets of keypoints. This approach is fully differentiable and thus allows end-to-end training by measuring the error of the predicted transformation with respect to the ground truth relative pose.
}

\subsection{Loss Functions}
\label{sec:loss}

Our total loss function consists of a weighted sum of the triplet loss $\mathcal{L}_\mathit{Tri}$ for loop closure detection as well as a geometric loss $\mathcal{L}_\mathit{Geo}$ and the newly proposed panoptic loss $\mathcal{L}_\mathit{Pan}$ for point cloud registration. The following paragraphs describe these losses in greater detail.


{\parskip=5pt
\noindent\textit{Triplet Loss:} 
For the loop closure detection task, we use the triplet loss.
It enforces a small distance between the descriptors of an anchor point cloud and a positive point cloud, i.e., a loop closure LiDAR scan while increasing the distance between the descriptors of the anchor and a negative point cloud, i.e., a LiDAR scan taken at a different place.
\begin{equation}
    \label{eq:loss_triplet}
    \mathcal{L}_\mathit{Tri} = \max \left\{ d( \mathbf{D}^\text{a}, \mathbf{D}^\text{p} ) - d( \mathbf{D}^\text{a}, \mathbf{D}^\text{n} ) + m , 0 \right\},
\end{equation}
where the descriptors of the anchor, the positive, and the negative sample are denoted by $\mathbf{D}^\text{a}$, $\mathbf{D}^\text{p}$, and $\mathbf{D}^\text{n}$, respectively. $d(\cdot)$ is a given distance function and $m$ refers to the desired separation margin.
}


{\parskip=5pt
\noindent\textit{Geometric Loss:}
We formulate our geometric loss $\mathcal{L}_\mathit{Geo}$ as a sum of a pose loss $\mathcal{L}_\mathit{Pos}$ and an auxiliary matching loss $\mathcal{L}_\mathit{Mat}$.
For the pose loss, we compare the predicted relative transformation $\widehat{H}^p_a$ from the anchor to the positive sample with the ground truth transformation $H^p_a$ by applying both to the coordinates of the same sampled point cloud $\mathbf{Q}^\text{a}$. Then we compute the mean absolute error in the Euclidean space.
\begin{equation}
    \label{eq:loss_pose}
    \mathcal{L}_\mathit{Pos} = \operatorname{mean} \left( \operatorname{abs} \left( \widehat{H}^p_a \cdot \mathbf{Q}^\text{a} - H^p_a \cdot \mathbf{Q}^\text{a} \right) \right)
\end{equation}

We further evaluate the geometric correspondence between the sampled anchor $\mathbf{Q}^\text{a}$ and positive points $\mathbf{Q}^\text{p}$ leveraging the predicted matching matrix $\widehat{M}^a_p$. In detail, we transform the anchor points with the ground truth transformation $H^a_p$ and project the positive sample with $\widehat{M}^a_p$.
\begin{equation}
    \label{eq:loss_match}
    \mathcal{L}_\mathit{Mat} = \operatorname{mean} \left( \operatorname{abs} \left( H^p_a \cdot \mathbf{Q}^\text{a} - \widehat{M}^a_p \cdot \mathbf{Q}^\text{p} \right) \right)
\end{equation}
}


\begin{figure}[t]
    \centering
    \captionsetup[subfigure]{justification=centering}
    \subfloat[Object matching]{%
        \includegraphics[width=.485\linewidth]{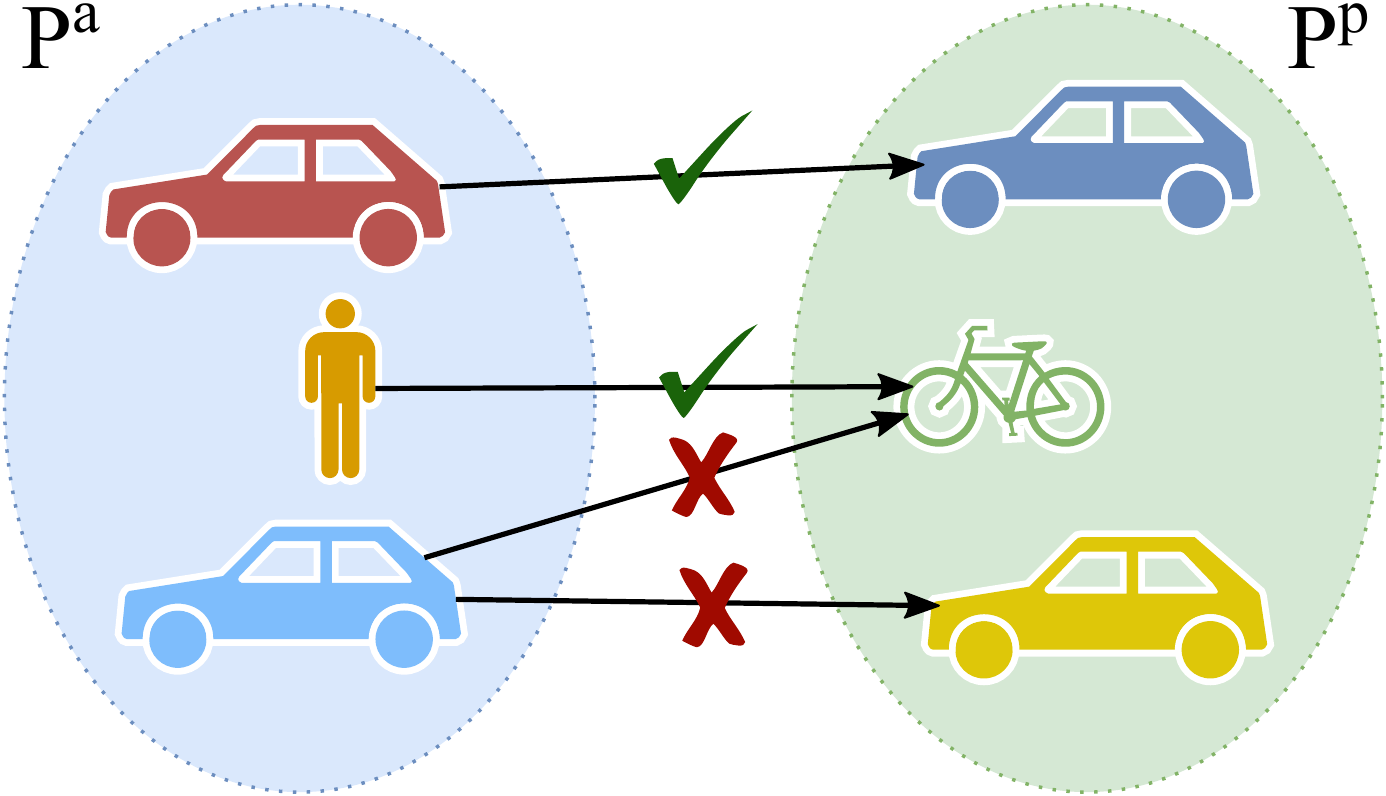}}
    \hfill
    \subfloat[Graph representation]{%
        \includegraphics[width=.485\linewidth]{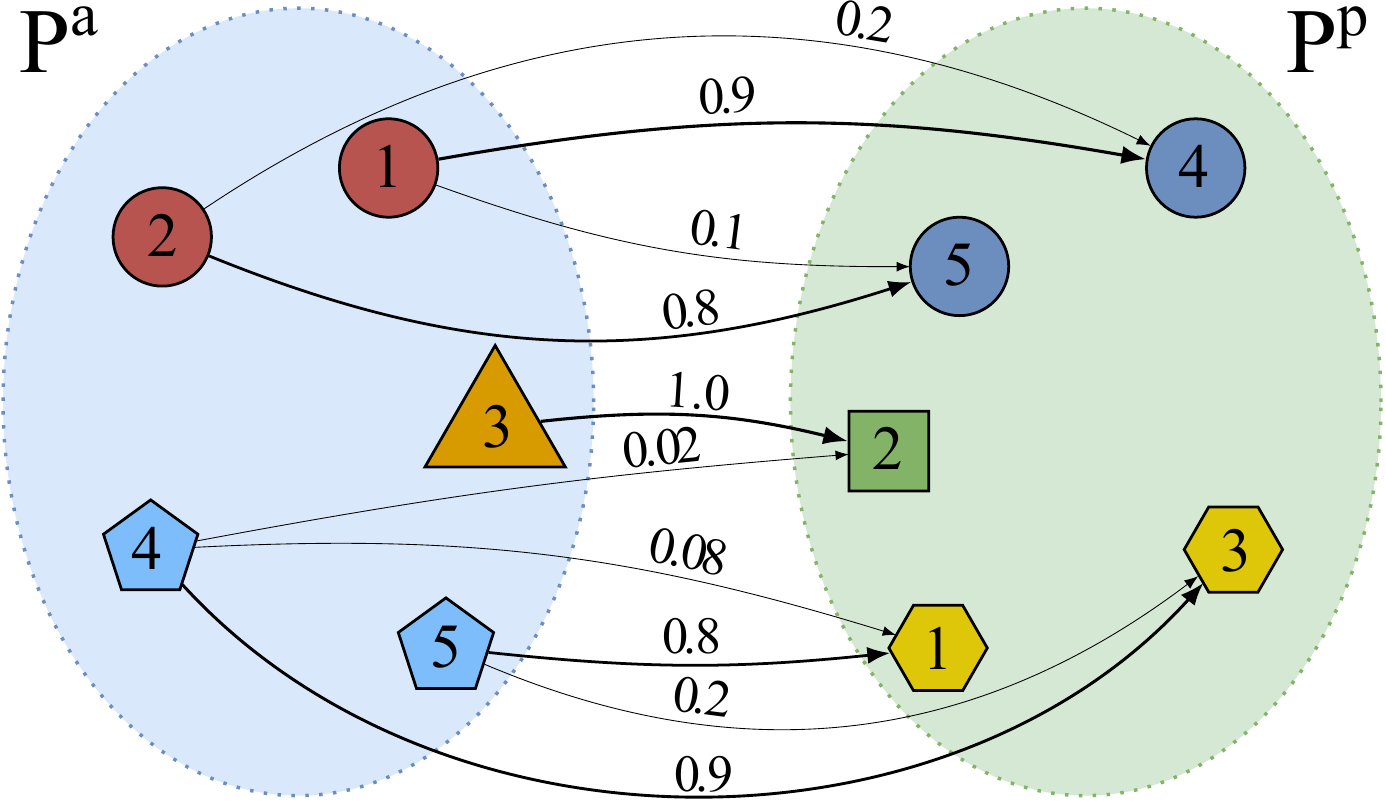}}
    \caption{The multi-matched object loss penalizes matching an object in the anchor point cloud to multiple objects in the positive sample. Unlike the semantic misclassification losses, the multi-matched object loss does not consider the semantic class, as depicted in (a). By exploiting a graph representation shown in (b) of the point cloud, it enforces that all points of the same object are matched to points of another object.}
    \label{fig:mmo_loss}
    \vspace{-.5cm}
\end{figure}


{\parskip=5pt
\noindent\textit{Panoptic Loss:}
In addition to the geometric point correspondences, we propose to leverage panoptic information to register two point clouds. In detail, we formulate a novel panoptic loss $\mathcal{L}_\mathit{Pan}$ as the sum of semantic misclassification losses $\mathcal{L}_\mathit{Sem}$ and $\mathcal{L}_\mathit{Mes}$ as well as a multi-matched object loss $\mathcal{L}_\mathit{Mmo}$.

We treat the matching process as a classification task, where the projected positive points are assigned a semantic class. While a cross-entropy loss is commonly used in classification problems, due to the fact that the proposed class logits are not the output of either a logistic or softmax activation, we empirically found that a mean absolute error resulted in a more stable training process. First, we use the semantic labels to construct one-hot encoded matrices $\mathbf{K}^\text{a}$ and $\mathbf{K}^\text{p}$ for the anchor and positive samples, respectively.
Using the predicted matching matrix $\widehat{M}^a_p$, we define the semantic loss as
\begin{equation}
    \label{eq:loss_semantic}
    \mathcal{L}_\mathit{Sem} = \operatorname{mean} \left( \operatorname{abs} \left( \mathbf{K}^\text{a} - \widehat{M}^a_p \cdot \mathbf{K}^\text{p} \right) \right).
\end{equation}

Additionally, to allow flexibility in the semantic misclassification, we define a mapping from the semantic class labels to a set of super-classes, e.g., both \textit{car} and \textit{truck} belong to the \textit{vehicle} class. Further details can be found in \refsec{sec:implementation}. Analogously to the semantic loss, we construct one-hot encoded matrices $\mathbf{J}^a$ and $\mathbf{J}^p$ and define the meta-semantic loss as
\begin{equation}
	\label{eq:loss_metasemantic}
    \mathcal{L}_\textit{Mes} = \operatorname{mean} \left( \operatorname{abs} \left( \mathbf{J}^\text{a} - \widehat{M}^a_p \cdot \mathbf{J}^\text{p} \right) \right).
\end{equation}

In our novel multi-matched object loss, we further exploit the instance labels to encourage the network to match entire objects consistently from one point cloud to the other. This is done by penalizing matches of points from a single object in the anchor to multiple objects in the positive sample. Unlike the previously introduced semantic misclassification losses, the multi-matched object loss does not consider the semantic class of objects, as depicted in \reffig{fig:mmo_loss}~(a).

Since instance labels may not be consistent throughout a driving sequence, it is not feasible to purely rely on the IDs. Therefore, we construct adjacency matrices $\mathbf{O}^\text{a}$ and $\mathbf{O}^\text{p}$ of a graph representation of the point clouds, where nodes represent points and edges connect points of the same instances of a semantic class. The predicted matching matrices $\widehat{M}^a_p$ and $\widehat{M}^p_a$ can then be considered as weighted, directed, bipartite graphs between the two sets of points (see \reffig{fig:mmo_loss}~(b)). Finally, we formulate the multi-matched object loss as
\begin{equation}
    \label{eq:loss_object}
    \mathcal{L}_\mathit{Mmo} = \operatorname{mean} \left( \left(1 - \mathbf{O}^\text{a} \right) \odot \left( \widehat{M}^a_p \cdot \mathbf{O}^\text{p} \cdot \widehat{M}^p_a \right) \right),
\end{equation}
where $\odot$ denotes the element-wise multiplication.
}


{\parskip=5pt
\noindent\textit{Reverse Losses:}
Finally, we add a second instance of the registration module that processes the swapped source~\(s\) and target~\(t\) inputs and predicts the inverse relative transformation.
Both the geometric and the panoptic losses can be reformulated accordingly. The total loss is then formulated by averaging the results of both the original and the reverse versions.
}

\section{Experimental Evaluation}

In this section, we evaluate our proposed PADLoC architecture with respect to multiple handcrafted and learning-based methods. We perform several experiments and present both the loop closure detection and the point cloud registration results. Finally, we evaluate the design choices in PADLoC by performing multiple ablation studies and provide a brief efficiency analysis.


\begin{table*}[t]
\footnotesize
\centering
\caption{Comparison of loop closure detection and point cloud registration performance}
\label{tab:generalization}
\vspace*{-.2cm}
\setlength{\tabcolsep}{3pt}
\begin{threeparttable}
    \begin{tabular}{ c l  c c c c c  c c c c c  c c c c c }
        \toprule
        & & \multicolumn{5}{c}{KITTI Seq. 08~\cite{geiger2012are}} & \multicolumn{5}{c}{Ford Seq. 01~\cite{pandey2011ford}} & \multicolumn{5}{c}{Freiburg \textit{(in-house)}} \\
        \cmidrule(lr){3-7} \cmidrule(lr){8-12} \cmidrule(lr){13-17}
        & Method & AP & Max-F1 & EP & $r_{\mathit{err}}$ [\degree] & $t_{\mathit{err}}$ [m] & AP & Max-F1 & EP & $r_{\mathit{err}}$ [\degree] & $t_{\mathit{err}}$ [m] & AP & Max-F1 & EP & $r_{\mathit{err}}$ [\degree] & $t_{\mathit{err}}$ [m] \\
        \midrule
        \multirow{6}{*}{\begin{sideways} \scriptsize Handcrafted \end{sideways}}
        & M2DP~\cite{He2016} & 0.05 & 0.10 & 0.00 & --- & --- & 0.89 & 0.88 & 0.89 & --- & --- & 0.71 & 0.68 & 0.74 & --- & --- \\
        & Scan Context\textsuperscript{\asterisk}~\cite{kim2022scan} & 0.65 & 0.62 & 0.00 & 3.11 & --- & \underline{0.97} & \textbf{0.95} & \underline{0.94} & 16.68 & --- & 0.81 & \textbf{0.79} & \textbf{0.82} & 52.70 & --- \\
        & LiDAR-Iris\textsuperscript{\asterisk}~\cite{wang2020lidar} & 0.64 & 0.62 & \textbf{0.71} & \underline{1.84} & --- & 0.90 & 0.64 & 0.50 & \underline{1.66} & --- & 0.81 & \underline{0.78} & \textbf{0.82} & 46.24 & --- \\
        & ISC\textsuperscript{\asterisk}~\cite{wang2020intensity} & 0.31 & 0.32 & \underline{0.55} & 6.27 & --- & 0.62 & 0.70 & 0.00 & 6.15 & --- & 0.82 & 0.75 & \underline{0.79} & 51.02 & --- \\
        & ICP (pt2pt)~\cite{zhang1994iterative} & --- & --- & --- & 160.63 & 2.41 & --- & --- & --- & 9.56 & 2.79 & --- & --- & --- & 89.43 & 2.37 \\
        & ICP (pt2pl)~\cite{zhang1994iterative} & --- & --- & --- & 160.73 & 2.49 & --- & --- & --- & 9.16 & 2.62 & --- & --- & --- & 89.25 & 2.25 \\
        \midrule
        \multirow{5}{*}{\begin{sideways} \scriptsize Learning \end{sideways}}
        & DCP~\cite{wang2019deep} & --- & --- & --- & 46.06 & 2.59 & --- & --- & --- & 12.14 & 3.42 & --- & --- & --- & 45.70 & 2.30 \\
        & SGPR~\cite{kong2020semantic} & 0.06 & 0.13 & 0.00 & --- & --- & 0.11 & 0.27 & 0.01 & --- & --- & 0.15 & 0.31 & 0.05 & --- & --- \\
        & OverlapNet\textsuperscript{\asterisk}~\cite{chen2020overlapnet} & 0.32 & 0.37 & 0.50 & 65.45 & --- & 0.79 & 0.81 & 0.84 & 9.44 & --- & 0.76 & 0.72 & 0.76 & 70.91 & --- \\
        & LCDNet~\cite{cattaneo2022lcdnet} & \underline{0.76} & \underline{0.74} & 0.50 & \textbf{0.37} & \underline{0.19} & \underline{0.97} & \underline{0.93} & 0.72 & 1.82 & \underline{1.44} & \textbf{0.84} & 0.73 & 0.71 & \underline{10.08} & \textbf{0.91} \\
        & PADLoC (ours) & \textbf{0.81} & \textbf{0.78} & 0.51 & \textbf{0.37} & \textbf{0.16} & \textbf{0.98} & 0.85 & \textbf{0.95} & \textbf{1.50} & \textbf{1.33} & \underline{0.83} & 0.74 & 0.74 & \textbf{9.30} & \underline{1.41} \\
        \bottomrule
    \end{tabular}
    Comparison of the average precision (AP), the maximum F1 score, and the extended precision (EP)~\cite{ferrarini2020exploring} for loop closure detection as well as rotation error $r_{\mathit{err}}$ and translation error $t_{\mathit{err}}$ for point cloud registration of PADLoC with previous methods.
    All learning-based models are trained on the KITTI odometry benchmark dataset. PADLoC uses panoptic annotations from the SemanticKITTI dataset.
    Methods denoted with \textsuperscript{\asterisk} only estimate the yaw between two point clouds instead of a full 6-DoF transformation.
    Bold and underlined values denote the best and second best scores, respectively.
\end{threeparttable}
\vspace*{-.5cm}
\end{table*}


\subsection{Implementation Details}
\label{sec:implementation}

We perform experiments on two publicly available autonomous driving datasets, namely the KITTI odometry benchmark~\cite{geiger2012are} and the Ford campus vision and LiDAR dataset~\cite{pandey2011ford}. Additionally, we also present results on a more challenging in-house dataset recorded in Freiburg, Germany. For training PADLoC, we leverage the ground truth panoptic annotations from the SemanticKITTI dataset~\cite{behley2019semantickitti}. If not specified otherwise, we train all learning-based models on sequences \{00, 05, 06, 07, 09\} of KITTI and evaluate on sequence 08. For the results on the Ford and Freiburg datasets presented in \reftab{tab:generalization}, we do not retrain the methods but use the weights trained on KITTI. Unless otherwise specified, we use $n=4096$ keypoints, set the feature size to $f=640$, the descriptor length to $g=256$, and the number of clusters $k=64$. To improve the invariance of the model with respect to the inputs' position and orientation, we augment the data during training by applying a random rigid transformation to the input point clouds with a uniform translation of \SI{+-1.5}{\meter} in the \(x\) and \(y\) axes and \SI{+-0.25}{\meter} along \(z\), and a uniform rotation of \SI{+-3}{\degree} for the roll and pitch angles and \SI{+-180}{\degree} for the yaw. We train all our models on a server with 4 NVIDIA RTX A6000 GPUs for 150 epochs with a batch size of $b=8$. We use the Adam optimizer with an initial learning rate of $\lambda=0.004$, halved after epochs 40 and 80, and with a weight decay of $5\times10^{-6}$. 

The total loss function is computed as a weighted sum of the components described in \refsec{sec:loss}, with weights $w_{\mathit{Tri}}=1.0$, $w_{\mathit{Pos}}=1.0$, $w_{\mathit{Mat}}=0.05$, $w_{\mathit{Sem}}=0.125$, $w_{\mathit{Mes}}=0.5$, and $w_{\mathit{Mmo}}=10.0$. We use a triplet margin of $m=0.5$ and the L2 distance as the distance function in \refeqn{eq:loss_triplet}. For the semantic super-classes, we follow the definitions of Cityscapes~\cite{cordts2016the} and group the semantic labels into \textit{flat}, \textit{human}, \textit{vehicle}, \textit{construction}, \textit{object}, \textit{nature}, and \textit{void}. Based on the ablation study presented in \refsec{ssec:exp_ablation_studies}, we use the Berger-Parker index to compute the confidence weights.

\subsection{Loop Closure Detection}
\label{ssec:exp_loop_closure_detection}

To evaluate the loop closure detection performance, we compare PADLoC with the handcrafted methods M2DP~\cite{He2016}, Intensity Scan Context~(ISC)~\cite{wang2020intensity}, Scan Context~\cite{kim2022scan}, and LiDAR-Iris~\cite{wang2020lidar}, as well as with the learning-based approaches LCDNet~\cite{cattaneo2022lcdnet}, OverlapNet~\cite{chen2020overlapnet}, Deep Closest Point~(DCP)~\cite{wang2019deep}, and SGPR~\cite{kong2020semantic}, which uses panoptic information also during inference. For DCP, we combine the feature extraction module of PADLoC with a full transformer-based matching module based on the authors' code release. For the other methods, we directly use the official code published by the respective authors. To compute the results on OverlapNet and SGPR, we download the model weights provided on the project website that are trained on KITTI.
Since SGPR requires panoptic labels during inference time, we use predictions by RangeNet++~\cite{milioto2019rangenet} combined with point clustering to obtain instances.

When evaluating PADLoC, we generate a descriptor $\mathbf{D}_i$ for every scan $i$ in a sequence and compute its similarity with that of all frames prior to the 50 previous scans. If a scan $j$ with the closest descriptor to that of scan $i$ has a similarity higher than a threshold $\tau$, then the pair $(i, j)$ is considered to form a loop closure. If the distance between the two ground truth poses is within \SI{4}{\meter}/\SI{10}{\meter}/\SI{20}{\meter} for the KITTI/Ford/Freiburg dataset, then it is considered a true positive. Otherwise, it is considered a false positive. Conversely, if the pose distance is within \SI{4}{\meter}/\SI{10}{\meter}/\SI{20}{\meter}, but the similarity between the descriptors is below the threshold $\tau$, then we regard it as a false negative. By changing the value of $\tau$, we obtain precision-recall pairs that are then used to compute the average precision (AP).

\begin{figure*}[t]
\footnotesize
    \centering
    \includegraphics[width=.95\linewidth]{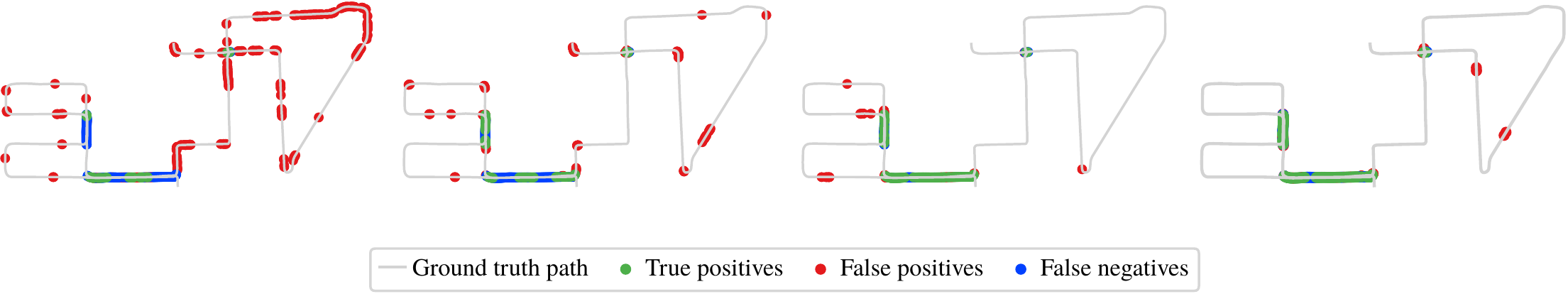}
    \begin{minipage}[t]{.245\textwidth}
        \vspace{-.1cm}
        \centering
        (a) SGPR~\cite{kong2020semantic}
    \end{minipage}%
    \hfill
    \begin{minipage}[t]{.245\textwidth}
        \vspace{-.1cm}
        \centering
        (b) OverlapNet~\cite{chen2020overlapnet}
    \end{minipage}%
    \hfill
    \begin{minipage}[t]{.245\textwidth}
        \vspace{-.1cm}
        \centering
        (c) LCDNet~\cite{cattaneo2022lcdnet}
    \end{minipage}
    \hfill
    \begin{minipage}[t]{.245\textwidth}
        \vspace{-.1cm}
        \centering
        (d) PADLoC (ours)
    \end{minipage}
    \vspace{-.1cm}
    \caption{Qualitative loop closure detection results on KITTI sequence 08 of the learning-based methods. The ground truth path corresponds to true negatives. While LCDNet reduces both false positives and false negatives compared to OverlapNet, the proposed PADLoC further decreases false positives.}
    \label{fig:lc_path}
    \vspace*{-.5cm}
\end{figure*}

In \reftab{tab:generalization}, we report the AP, the maximum F1 score, and the extended precision~(EP)~\cite{ferrarini2020exploring} of PADLoC and the aforementioned baseline methods. Notably, PADLoC achieves the highest AP and \mbox{Max-F1} score across the entire board for the evaluation sequences of KITTI and the highest AP and EP on Ford. On our in-house Freiburg dataset, PADLoC yields the highest Max-F1 score as well as the second best AP and EP compared to the other learning-based approaches. Although the proposed transformer-based registration head and the panoptic losses do not directly influence the loop closure detection module, by sharing the same feature extractor between the two branches and jointly training the two tasks, PADLoC learns a better feature representation improving the loop closure detection performance compared to LCDNet, which achieved the second best AP on both KITTI and Ford. Qualitative results of these methods on the KITTI dataset are visualized in \reffig{fig:lc_path}. Compared to OverlapNet, both LCDNet and PADLoC correctly detect a higher number of loop closures, whereas PADLoC is able to further reduce the number of false positives. While the learning-based methods LCDNet and PADLoC outperform all handcrafted methods when evaluated on the same domain as used for training, this gap vanishes on Ford and Freiburg. Here, these methods perform on par with the best handcrafted approach Scan~Context.


\subsection{Point Cloud Registration}
\label{ssec:experiment_registration}

To evaluate the point cloud registration performance, we compare PADLoC with the same handcrafted and learning-based methods described in \refsec{ssec:exp_loop_closure_detection}, except for M2DP and SGPR that do not perform point cloud registration. Since the handcrafted methods only estimate the yaw between two point clouds instead of the full 6-DoF transformation, we additionally compare with the Iterative Closest Point algorithm~(ICP)~\cite{zhang1994iterative}, using both point-to-point and point-to-plane distances. Following the standard experimental setup~\cite{cattaneo2022lcdnet}, for LCDNet, DCP, and PADLoC, we perform point cloud registration with RANSAC using the extracted features before the respective matching layers.

As a measure of registration accuracy, we compute the rotation error $r_\mathit{err}$ in degrees and the translation error $t_\mathit{err}$ in meters of all positive pairs. We then average the errors over the entire sequence and present the results in \reftab{tab:generalization}. We observe that PADLoC yields the smallest rotation error compared to all the handcrafted and learning-based methods on each of the evaluation sequences in the datasets. Additionally, it yields the smallest translation error on both the KITTI and Ford datasets, as well as the second lowest translation error on our in-house Freiburg dataset. LCDNet achieves the second best performance in most evaluations while achieving the lowest translation error on the Freiburg dataset. This result shows that while the feature extraction architecture and the training scheme play an important role, leveraging the cross-modal attention matrices from the transformer architecture and the panoptic information during training further improves the point cloud registration performance. While LiDAR-Iris achieves the lowest rotation error across all the handcrafted methods, it only estimates the yaw angle instead of the full 6-DoF transformation.\looseness=-1


\begin{table}
\footnotesize
\centering
\caption{Ablation study on confidence weights}
\label{tab:abl_conf}
\vspace{-.2cm}
\setlength{\tabcolsep}{7.5pt}
\begin{threeparttable}
    \begin{tabular}{ l c c c }
        \toprule
        Method & AP $\uparrow$ & $r_{\mathit{err}}$ [\degree] $\downarrow$ & $t_{\mathit{err}}$ [m] $\downarrow$ \\
        \midrule
        Uniform         & 0.73 & 4.63 & 3.76 \\
        Column sum      & 0.76 & 6.34 & 3.62 \\
        Shannon         & 0.50 & 21.86 & 3.99 \\
        Hill (r=2)      & \textbf{0.89} & 2.45 & 2.00 \\
        Hill (r=4)      & 0.84 & 2.47 & 2.12 \\
        Berger-Parker   & 0.81 & \textbf{2.35} & \textbf{1.43} \\
        \bottomrule
    \end{tabular}
    \vspace{1pt}
    Average precision (AP) of loop closure detection as well as the mean error of point cloud registration, evaluated on KITTI sequence 08 for different weightings used in SVD.
\end{threeparttable}
\vspace{-.5cm}
\end{table}

\subsection{Ablation Studies}
\label{ssec:exp_ablation_studies}

In this section, we present ablation studies to analyze the major design choices of PADLoC. As the RANSAC-based point cloud registration described in \refsec{ssec:experiment_registration} is applied only during inference and does not impact the training stage, the experiments in this section do not exploit RANSAC.


{\parskip=3pt
\noindent\textit{Confidence Weighting:}
We investigate the effect of different weighting schemes on the performance of both loop closure detection and point cloud registration tasks. In \reftab{tab:abl_conf}, we present the average precision (AP) as well as the registration errors $r_\mathit{err}$ and $t_\mathit{err}$ for the six weighting methods. In particular, uniform weights corresponding to unweighted SVD, column sum representing the method used in LCDNet~\cite{cattaneo2022lcdnet}, where weights are the sums along the columns of the matching matrix, and the diversity metrics from \refsec{sec:matching}, i.e., the Shannon Entropy, the order-$r$ Hill number with $r \in \{2, 4\}$, and the Berger-Parker index. We observe that both the Hill numbers and the Berger-Parker index outperform the other confidence weighting methods. Due to the substantially smaller translation error of the Berger-Parker index, improving the registration by more than \SI{0.5}{\meter}, we use this method in our final design.
}


{\parskip=3pt
\noindent\textit{Effect of Losses:}
To demonstrate the efficacy of our proposed panoptic loss $\mathcal{L}_\mathit{Pan}$ and the impact of formulating all losses in a bidirectional manner ($\mathcal{L}_\mathit{Rev}$), we consecutively add them to the original geometric loss $\mathcal{L}_\mathit{Geo}$. We present the results for both the loop closure detection and point cloud registration tasks in \reftab{tab:abl_losses}. We observe that adding the proposed panoptic losses increases the average loop closure detection precision by further constraining which points can be matched together based on their semantic and instance labels. Furthermore, by including the second matching and registration head, along with its corresponding reverse losses as illustrated in the bottom row, the added bidirectional consistency constraint yields the highest AP and the smallest registration errors. 
}

\begin{table}
\setlength{\tabcolsep}{4pt}
\footnotesize
\centering
\caption{Influence of the loss functions}
\label{tab:abl_losses}
\vspace{-.2cm}
\begin{threeparttable}
    \begin{tabular}{ c c c c c c }
        \toprule
        $\mathcal{L}_\mathit{Geo}$ & $\mathcal{L}_\mathit{Pan}$ & $\mathcal{L}_\mathit{Rev}$ & AP $\uparrow$ & $r_{\mathit{err}}$ [\degree] $\downarrow$ & $t_{\mathit{err}}$ [m] $\downarrow$ \\
        \midrule
        \cmark &        &        & 0.70 & 3.09 & 1.62 \\
        \cmark & \cmark &        & 0.78 & 3.36 & 1.71 \\
        \cmark & \cmark & \cmark & \textbf{0.81} & \textbf{2.35} & \textbf{1.43} \\
        \bottomrule
    \end{tabular}
    \vspace{1pt}
    Average precision (AP) of loop closure detection and the mean error of point cloud registration, evaluated on KITTI sequence 08 for the different loss functions.
\end{threeparttable}
\vspace{-.6cm}
\end{table}


\subsection{Efficiency Analysis}
\label{ssec:exp_efficiency_analysis}

We evaluate the memory footprint and inference time of our method on an NVIDIA RTX 3090 GPU. PADLoC requires \SI{3.4}{GB}. Compared to the full transformer-based matching module of DCP that requires \SI{10.3}{GB}, the memory footprint of PADLoC is only one-third, showing its lower complexity.

On average, PADLoC needs \SI{10}{\milli\second} for pre-processing a single point cloud. The shared feature extraction step consumes \SI{167}{\milli\second}. Computing the global descriptor used for loop closure detection takes \SI{0.1}{\milli\second} per point cloud. Finally, one forward pass of the registration and matching module to compute the transform between two point clouds takes \SI{14}{\milli\second}.

\section{Conclusion}

In this paper, we proposed the novel PADLoC architecture for LiDAR-based joint loop closure detection and point cloud registration. PADLoC is composed of a common feature extractor, a global descriptor as well as a transformer-based registration and matching module. Unlike previous approaches, we feed different inputs as value, query, and key to the transformer encoder exploiting its internal structure. We further introduced a new loss function that leverages ground truth panoptic annotations by penalizing matching points from different semantic classes as well as across multiple objects and validated its positive impact. Through extensive experimental evaluations, we demonstrated the efficacy of PADLoC compared to both handcrafted and learning-based methods. Future work will focus on exploiting panoptic information in an online manner and applying the matching approach of PADLoC to point cloud registration tasks in other domains.


\footnotesize
\bibliographystyle{IEEEtran}
\bibliography{references.bib}


\end{document}